\documentclass{article}

\usepackage[nonatbib,preprint]{nips_2018}

\usepackage[utf8]{inputenc} % allow utf-8 input
\usepackage[T1]{fontenc}    % use 8-bit T1 fonts
\usepackage{hyperref}       % hyperlinks
\usepackage{url}            % simple URL typesetting
\usepackage{booktabs}       % professional-quality tables
\usepackage{amsfonts}       % blackboard math symbols
\usepackage{nicefrac}       % compact symbols for 1/2, etc.
\usepackage{microtype}      % microtypography

\usepackage{cite}
\usepackage{varwidth}
\usepackage{float}
\usepackage{graphicx}
\usepackage{caption}
\usepackage{amsmath}
\usepackage{multirow}
\usepackage[ruled,vlined]{algorithm2e}
\usepackage{cleveref}

\title{A Methodology for Automatic Selection of Activation Functions to Design Hybrid Deep Neural Networks}

\author{
  Alberto Marchisio\\
  Vienna University of Technology\\
  Vienna, Austria \\
  \texttt{alberto.marchisio@tuwien.ac.at} \\
  \And
  Muhammad Abdullah Hanif \\
  Vienna University of Technology \\
  Vienna, Austria \\
  \texttt{muhammad.hanif@tuwien.ac.at} \\
  \And
  Semeen Rehman \\
  Vienna University of Technology \\
  Vienna, Austria \\
  \texttt{semeen.rehman@tuwien.ac.at} \\
  \And
  Maurizio Martina \\
  Politecnico di Torino \\
  Turin, Itlay \\
  \texttt{maurizio.martina@polito.it} \\
  \And
  Muhammad Shafique \\
  Vienna University of Technology \\
  Vienna, Austria \\
  \texttt{muhammad.shafique@tuwien.ac.at} \\
}

\begin{document}
\maketitle

\begin{abstract}
Activation functions influence behavior and performance of DNNs. Nonlinear activation functions, like Rectified Linear Units (ReLU), Exponential Linear Units (ELU) and Scaled Exponential Linear Units (SELU), outperform the linear counterparts. However, selecting an appropriate activation function is a challenging problem, as it affects the accuracy and the complexity of the given DNN. In this paper, we propose a novel methodology to automatically select the best-possible activation function for each layer of a given DNN, such that the overall DNN accuracy, compared to considering only one type of activation function for the whole DNN, is improved. However, an associated scientific challenge in exploring all the different configurations of activation functions would be time and resource-consuming. Towards this, our methodology identifies the Evaluation Points during learning to evaluate the accuracy in an intermediate step of training and to perform early termination by checking the accuracy gradient of the learning curve. This helps in significantly reducing the exploration time during training. Moreover, our methodology selects, for each layer, the dropout rate that optimizes the accuracy. Experiments show that we are able to achieve on average 7 \% to 15 \% Relative Error Reduction on MNIST, CIFAR-10 and CIFAR-100 benchmarks, with limited performance and power penalty on GPUs.

\end{abstract}

\section{Introduction}
\label{sec:introduction}
\begin{figure}[h]
	\centering
	\includegraphics[width=\linewidth]{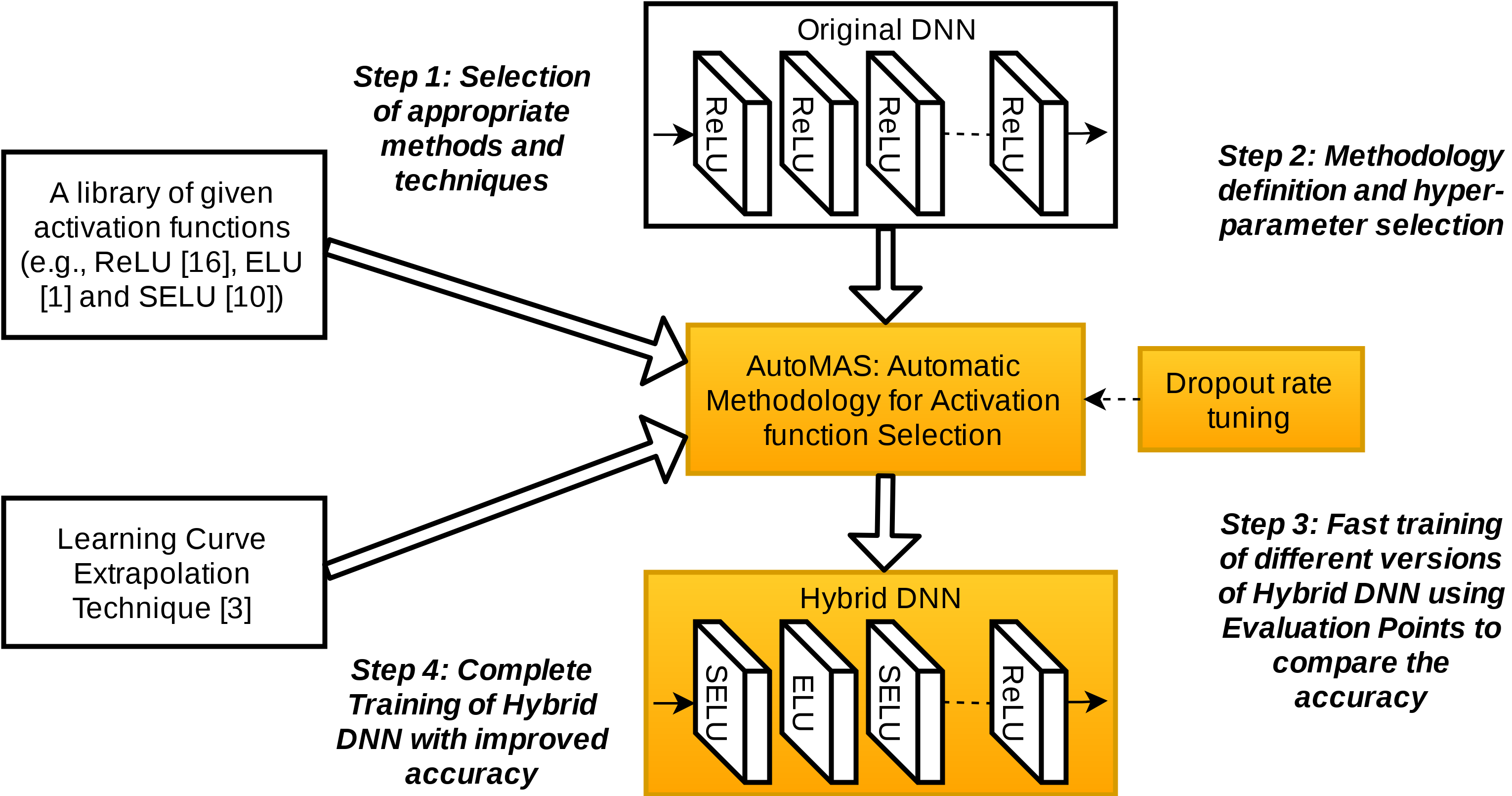}
	\caption{An Overview of our Methodology.}
	\label{fig:overview}
\end{figure}

Convolutional Neural Networks (CNN) are very popular among Artificial Intelligence applications, like computer vision \cite{ref:AlexNet}, speech recognition \cite{ref:speech_recognition} and natural language processing \cite{ref:natural_language}. In particular, they have emerged as the state-of-the-art in computer vision tasks \cite{ref:AlexNet,ref:VGGNet,ref:GoogleNet,ref:ResNet}. The key to their success is the significant improvement on GPUs' support for high performance and parallel computation, that allows Deep Neural Networks (DNN) to be trained in a reasonable amount of time.\\
DNNs usually require nonlinear activation functions. Rectified Linear Units (ReLU) \cite{ref:RELU} are widely used in state-of-the-art DNNs, because of their simplicity. Ioffe and Szegedy \cite{ref:Batch Normalization} proposed the Batch Normalization (BN) technique, whose usage improves the accuracy compared to original CNNs without BN. Recently, Klambauer et al. \cite{ref:SNN} introduced the Scaled Exponential Linear Unit (SELU) and demonstrated that, for a specific choice of parameters, it has self-normalizing properties, because it leads to zero mean and unit variance. Another important property of SELU functions, which is also applicable to Exponential Linear Units (ELU) \cite{ref:ELU}, is that the negative weights are continuously updated. This effect results in a wider learning ability of ELU and SELU compared to ReLU, that leads to an accuracy improvement.\\
\textbf{A key scientific question is:} \textit{Can we replace the ReLU layers of a given DNN by ELU or SELU layers to improve the DNN accuracy? If yes, how can we select which of the ReLU layers should be replaced by ELU or SELU layers, without significantly affecting the total training time?}\\
\textbf{Our novel contributions:} To address the above question, we propose a novel methodology that automatically explores different activation functions in each layer of the DNN and converges to select the model that leads to further accuracy improvements. Through this methodology, we can obtain a Hybrid DNN that can have ReLU, ELU or SELU activations in different layers, interchangeably. Since SELU functions require ``alpha'' dropout, we introduce the possibility to have dropout at each layer of the DNN, using standard dropout in case of ReLU and ELU and ``alpha'' dropout in case of SELU, to obtain a fair comparison. Our methodology allows to fine tune the dropout rate to achieve a higher accuracy compared to the original DNN. Instead of replacing activation functions in all the layers of the network, \textit{our methodology adds a degree of freedom in the DNN architecture, because different types of activation functions can be selected in different layers}. Since exploring all the configurations of activation functions extensively is an extremely compute-intensive task and may not be feasible in practice, \textit{our methodology evaluates different activation functions for each layer in an intermediate step of the training process, called the ``Evaluation Point''}. Such an Evaluation Point (EP) is automatically selected after a gradient-based analysis of the accuracy at each training epoch, i.e., after each forward and backward pass of all the training data through the DNN. We demonstrate (see \Cref{subsec:case_study}) that a comparison at the EP is effective, i.e., it produces similar results with respect to comparing the versions at the end of training. This optimization allows our methodology to reduce the exploration time during training by a factor varying from 4x to 7x, for different types of original DNNs. The final outcome of our methodology is a so-called Hybrid DNN that has the best configuration of activation functions and dropout rate, among all the possibilities, for each activation layer. Quantitatively, we achieve from 7 \% to 15 \% Error Rate Reduction in our Hybrid DNNs compared to the original versions. An overview of our novel contributions with inputs/outputs is illustrated in \Cref{fig:overview}.\\
\ \\
\textbf{Paper Organization:} \Cref{sec:related} discusses the related work. \Cref{sec:act_analysis} presents a quick analysis of the activation functions that we are using. \Cref{sec:methodology} presents our methodology and novel contributions. \Cref{sec:experiments} reports the experimental results on different benchmarks. \Cref{sec:conclusions} concludes the paper and summarizes the achievements.
\section{Related work}
\label{sec:related}
The ReLU activation function, introduced by Nair and Hinton \cite{ref:RELU}, has shown its great potentials in AlexNet \cite{ref:AlexNet} and other networks afterwards. Since it has zero derivative for negative inputs, the backpropagation error is blocked in those conditions. This is called the ``dead neuron problem'', because, once a neuron reaches this state, it will not escape and can be considered dead, since it cannot be updated. Many researchers proposed solutions for that problem. Maas et al. \cite{ref:Leaky RELU} suggested to use Leaky ReLU, where also the negative part of the activation function has a positive (linear) slope. Setting the appropriate value of the slope can be tricky, but He et al. \cite{ref:PRELU} showed a method to learn the slope automatically during backpropagation. Another important research direction is Exponential Linear Unit (ELU), proposed by Clevert et al. \cite{ref:ELU}. ELU is an activation function with exponential behavior in the negative part and linear in the positive one. Afterwards, Ioffe and Szegedy \cite{ref:Batch Normalization} implemented the Batch Normalization (BN) for ResNet \cite{ref:ResNet}, that contributes to its accuracy improvement over the previous state-of-the-art DNNs. Every layer using one of the activation functions described above may be associated to a BN layer to increase the accuracy of a given DNN. A recent work by Klambauer et al. \cite{ref:SNN} showed that Self-Normalizing Neural Networks (SNN) have the intrinsic property to automatically converge to zero mean and unit variance, without requiring explicit Batch Normalization. They propose to use Scaled Exponential Linear Units (SELU) as activation function. Several other novel activation functions can be generated automatically, as shown by the work of Ramachandran et al. \cite{ref:search_activation_function}. Selecting the appropriate activation function is not an easy task. Recent works by Harmon et al. \cite{ref:activation_ensembles} and Manessi et al. \cite{ref:learning_combinations} proposed to use a combination of different activation functions in the same layer, with connections learned during training. This approach, however, increases the memory footprint of the DNN by a significant factor, which is typically a critical parameter in real-world scenarios. Replacing each activation function throughout the complete network in a uniform way (Pinamonti \cite{ref:act_function_comparison}) shows that ELU and SELU outperform the other activation functions, but this is not an exhaustive search to find high accuracy improvements. To the best of our knowledge, a Hybrid DNN with different activation functions within different layers has not been explored. Our methodology has the potential to further improve the network accuracy, as we will demonstrate in \Cref{sec:experiments}. However, an appropriate selection of different activation functions in different layers demands a comprehensive evaluation with a huge trial-and-error effort if it is done in a naive way, thereby requiring an automatic methodology equipped with a fast evaluation strategy.\\
%Harmon et al. \cite{ref:activation_ensembles} proposed the Activation Ensembles, that allows to use multiple activation functions at each layer of the DNN, whose contributions are learned during backpropagation. Despite they obtain accuracy improvements with respect to the baseline network, the main challenge of this approach is that the complexity and storage requirements introduced for this scope make Activation Ensembles unfeasible to deploy in real-case scenarios, when the memory footprint of the DNN is a critical parameter.\\
\textit{In this paper, we propose an automatic methodology to systematically select different types of activation functions in different layers selectively, obtaining a Hybrid DNN.} The exploration time is significantly reduced with respect to the trial-and-error approach, because our methodology compares the versions in an intermediate training epoch, called Evaluation Point.
%The previous work by Domhan et al. \cite{ref:extrapolation_of_learning_curve} shows how to extrapolate useful information from the first epochs of the learning curve to optimize the hyper-parameters. We revisited this method to establish the Evaluation Point of our methodology.
%
\section{Activation Function Analysis}
\label{sec:act_analysis}
In this section, we briefly discuss, analyze and compare different activation functions (i.e., ReLU, ELU, and SELU) that are explored by our methodology. Other activation functions are not considered in this paper, but can easily be integrated in our methodology, as it takes a library of activation functions as an input (see \Cref{fig:overview}). For each of them and each first derivative, we define the respective analytic expressions and the behavior; see \Cref{fig:activation,fig:activation_derivative}. \Cref{subsec:dropout} describes the dropout method that fits with the different activations.\\
\begin{table}[h]
	\begin{varwidth}[b]{0.48\linewidth}
		\centering
		\includegraphics[width=\linewidth]{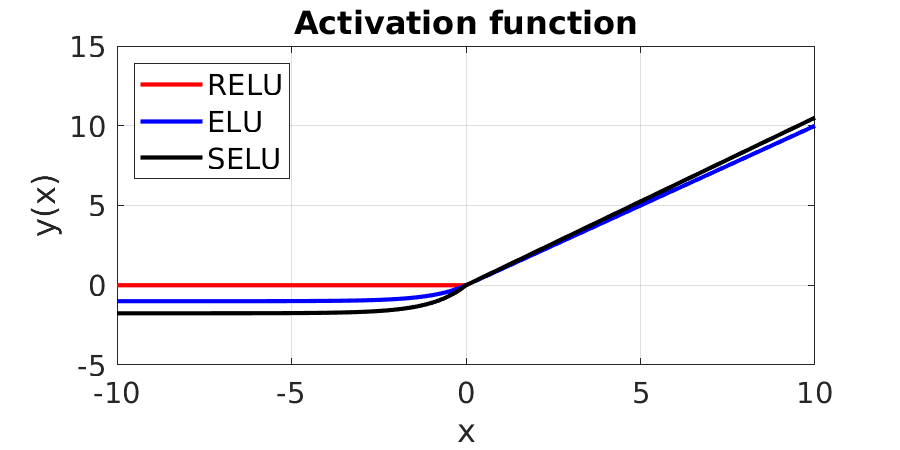}
		\captionof{figure}{Behavior of ReLU, ELU and SELU activation functions.}
		\label{fig:activation}
	\end{varwidth}
	\hfill
	\begin{varwidth}[b]{0.48\linewidth}
		\centering
		\includegraphics[width=\linewidth]{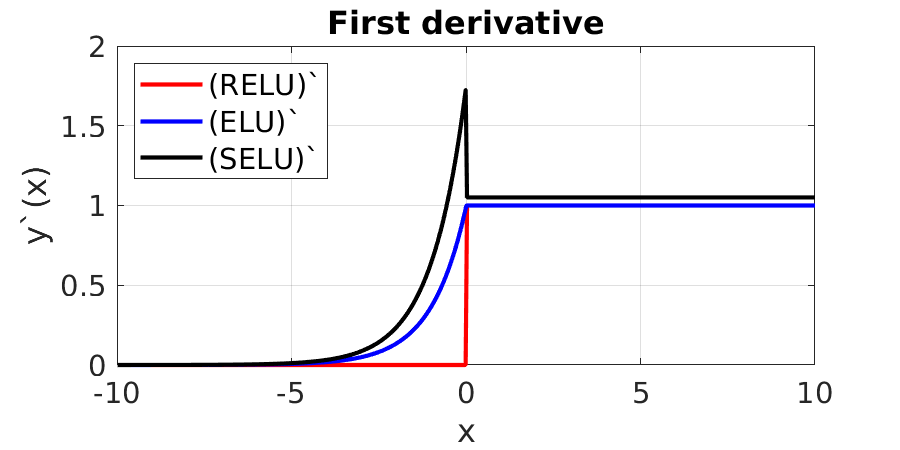}
		\captionof{figure}{Behavior of the first derivative of ReLU, ELU and SELU activation functions.}
		\label{fig:activation_derivative}
	\end{varwidth}
\end{table}
\subsection{ReLU}
\label{subsec:ReLU}
ReLU is a nonlinear activation function, expressed by \Cref{eq:RELU}. It is simple to implement and the computational effort is minimal. Its first derivative is expressed in \Cref{eq:RELU'}.
\begin{table}[H]
	\begin{varwidth}[b]{0.48\linewidth}
		\begin{equation}
			ReLU(x)=
			\begin{cases}
				0, & \textit{if\ \ } x \leq 0\\
				x, & \textit{if\ \ } x > 0
			\end{cases}
			\label{eq:RELU}
		\end{equation}
	\end{varwidth}
	\hfill
	\begin{varwidth}[b]{0.48\linewidth}
		\begin{equation}
			ReLU'(x)=
			\begin{cases}
				0, & \textit{if\ \ } x \leq 0\\
				1, & \textit{if\ \ } x > 0
			\end{cases}
			\label{eq:RELU'}
		\end{equation}
	\end{varwidth}
\end{table}
\subsection{ELU}
\label{subsec:ELU}
Compared to ReLU, ELU has an exponential behavior for negative inputs. It introduces a new parameter, $\alpha$, which can be considered as a new hyper-parameter of the network. Clevert et al. \cite{ref:ELU} propose to select $\alpha = 1$, hence we also use this value in our experiments. The analytic expression of ELU and its first derivative are reported in \Cref{eq:ELU,eq:ELU'}, respectively.
\begin{table}[H]
	\begin{varwidth}[b]{0.48\linewidth}
		\begin{equation}
		ELU(x)=
		\begin{cases}
		\alpha e^x - 1, & \textit{if\ \ } x \leq 0\\
		x, & \textit{if\ \ } x > 0
		\end{cases}
		\label{eq:ELU}
		\end{equation}
	\end{varwidth}
	\hfill
	\begin{varwidth}[b]{0.48\linewidth}
		\begin{equation}
		ELU'(x)=
		\begin{cases}
		\alpha e^x, & \textit{if\ \ } x \leq 0\\
		1, & \textit{if\ \ } x > 0
		\end{cases}
		\label{eq:ELU'}
		\end{equation}
	\end{varwidth}
\end{table}
\subsection{SELU}
\label{subsec:SELU}
SELU is relatively more complex than ELU, because it has another parameter, $\lambda$. Klambauer et al. \cite{ref:SNN} proposed to use $\alpha = 1.67326324$ and $\lambda = 1.050700987$. These values are used also in our experiments. SELU expression and its first derivative are reported in \Cref{eq:SELU,eq:SELU'}, respectively.
\begin{table}[H]
	\begin{varwidth}[b]{0.48\linewidth}
		\begin{equation}
		SELU(x) = \lambda
		\begin{cases}
		\alpha e^x - \alpha, & \textit{if\ \ } x \leq 0\\
		x, & \textit{if\ \ } x > 0
		\end{cases}
		\label{eq:SELU}
		\end{equation}
	\end{varwidth}
	\hfill
	\begin{varwidth}[b]{0.48\linewidth}
		\begin{equation}
		SELU'(x) = \lambda
		\begin{cases}
		\alpha e^x, & \textit{if\ \ } x \leq 0\\
		1, & \textit{if\ \ } x > 0
		\end{cases}
		\label{eq:SELU'}
		\end{equation}
	\end{varwidth}
\end{table}
\subsection{Dropout method}
\label{subsec:dropout}
The dropout technique has been introduced by Srivastava et al. \cite{ref:dropout}, in order to improve the regularization and to avoid the overfitting problem in DNNs. It is widely used in the most common state-of-the-art DNNs because of its regularizing property and simple applicability with ReLU activation functions. He et al. \cite{ref:PRELU} proposed a weight initialization method that is efficient for ReLU activations, because it limits the variance. Clevert et al. \cite{ref:ELU} applied the same initialization criteria with ELU activations. Kingma et al. \cite{ref:variational_dropout} analyzed how the variance changes when the dropout is applied. Klambauer et al. \cite{ref:SNN} revised it and proposed a new initialization method and a new dropout technique, specific for SELU. Weights are initialized in such a way that the mean $E(w_i)=0$ and the variance $Var(w_i)=1/n$, where $n$ is the number of inputs. This methods leads to the global variance (sum of all variances of each weight in the same layer) equal to 1. For example, each weight of a layer with 100 inputs should be initialized as a gaussian variable with zero mean and variance equal to 0.01.\\
Thus, such a new dropout method, called ``alpha'' dropout, sets dropped weights to $\alpha'$, where $\alpha' = \lim_{x \rightarrow - \infty} SELU(x) = - \lambda \alpha$. It is effective with SELU activations because it preserves mean and variance. Hence, in the following sections, we adopt standard dropout when applied to ReLU and ELU activations and "alpha dropout" when dealing with SELU.
\section{Our Novel Methodology}
\label{sec:methodology}
We propose a simple yet effective methodology to automatically select the activation functions for each layer of a given DNN as well as its associated dropout rate, based on the accuracy obtained at the Evaluation Point.
\subsection{Motivation and Key Features}
\label{subsec:motivations}
Activation function selection is a quite complex task and has a lot of implications on the performance and the accuracy of a given DNN. A simple selection process based on exploration requires extensive trial-and-error analysis to converge and yet it cannot guarantee to find a high-quality solution. Hence, our methodology, at the very first stage, focuses on an efficient way to extract useful information from the learning curve (i.e., the curve that describes the accuracy of the DNN as a function of the number of epochs) to obtain the Evaluation Point. Then, for each layer, we find the best combination (that produces the maximum test accuracy) of the activation function and the dropout rate. A layer-wise search is efficient for (1) improving the DNN accuracy and (2) not penalizing the computation efficiency, while using parallel processing and SIMD instructions in GPUs. Alternatively, our Hybrid DNN can easily be implemented and integrated in a hardware accelerator for Deep Learning Inference. Moreover, due to the Evaluation Point optimization, (3) we efficiently reduce the exploration time during the training process.
\subsection{Evaluation Point}
\label{subsec:evaluation_point}
The work by Domhan et al. \cite{ref:extrapolation_of_learning_curve} showed how to extrapolate useful information from the first epochs of the learning curve to optimize the hyper-parameters. We exploited this method to establish the concept of an Evaluation Point, which is computed during the first stage of our methodology. This allows us to perform early termination of the training process without significantly sacrificing the accuracy. Looking at the learning curve, we can identify a first region, where the accuracy (A) grows fast, and a second region, where the accuracy is almost flat. A re-parametrization that evaluates the gradient of such a curve, i.e., the so-called Accuracy Gradient (AG), allows us to define an analytical function that is able to find the Evaluation Points automatically. We define the AG as the average of the relative accuracy difference over a range (R) of epochs (E). It is expressed by \Cref{eq:AG}.
\begin{equation}
	AG(E) = \frac{\sum\limits_{i = E}^{E + R - 1} A(i) - \sum\limits_{j = E - R + 1}^{E} A(j)}{R \cdot A(E)}
	\label{eq:AG}
\end{equation}
The Evaluation Point (EP) is defined as the first epoch where the AG falls below 0.1 \%. Refer to \Cref{fig:accuracy_lenet,fig:accuracy_gradient} to see an example of the learning curve and the Accuracy Gradient to compute the Evaluation Point on the MNIST benchmark.
\subsection{In-depth view of Methodology}
\label{subsec:automas_methodology}
The essence of our methodology consists of training different versions of the DNN for the number of epochs specified by the EP. It is an iterative process, where, for each layer (except the last one, where the softmax activation function is typically used for classification purposes), we search the combination of an activation function and a dropout rate that optimizes the accuracy. We define the library of activation functions, composed by ReLU, ELU and SELU. For each activation function, we select the best dropout rate, according to what we call the ``smart'' search. Instead of not evaluating dropout, such a search explores the different values of the dropout rate, keeping track of the optimal dropout rate for the previous layer and moving step-by-step in intervals of [.1, .2, .5]. It moves one step further when the accuracy at the EP is increased with respect to the previous version and it changes the activation function otherwise. Once such a ``smart'' search is complete for the current layer, the current configuration is saved and we move on to analyzing the next layer. \Cref{alg:methodology} describes the flow of our methodology.
\begin{algorithm}[h]
	\caption{Our methodology. $L$ is the number of layers of the DNN; $F$ is the library of activation functions (i.e., F $\in$ \{ReLU, ELU, SELU\} in our case); $a_{f,l}$ is the accuracy achieved using the activation $f$ on the layer $l$; $d_{f,l}$ is the dropout rate of layer $l$ when using the activation function $f$; $f_l$ and $d_l$ are the activation function and the dropout rate of layer $l$, respectively.}
	\label{alg:methodology}
	1. Train the original DNN (with ReLU) completely;\\
	2. Compute the Evaluation Point;\\
	3. \For{$l=1$ \textbf{to} $L-1$}{
		$a_{0,l}^*=0$;\\
		$f^*=0$;\\
		\For{$f=1$ \textbf{to} $F$}{
			$a_{f,l}^*=0$;\\
			\While{$d_{f,l}$ is not optimized for accuracy}{
				Tune $d_{f,l}$;\\
				Train the DNN until the EP;\\
				Compute $a_{f,l}$;\\
				\If{$a_{f,l} > a_{f,l}^*$}{
					$a_{f,l}^*=a_{f,l}$;\\
					$d_{f,l}^*=d_{f,l}$;
				}
			}
			\If{$a_{f,l}^* > a_{f^*,l}^*$}{
				$f^*=f$;
			}
		}
		$f_l=f^*$;\\
		$d_l=d_{f^*,l}^*$;
	}
	4. Train the Hybrid DNN completely;
\end{algorithm}
\subsection{Hybrid DNN}
\label{subsec:hybrid_dnn}
Once we have processed every layer of the DNN, we train the resulting network, which we call Hybrid DNN, for the complete number of epochs (beyond the Evaluation Point). Such Hybrid DNN could have different activation functions in different layers, according to the selection criteria followed in \Cref{subsec:automas_methodology}, and has a better accuracy compared to the original version of a given DNN.
\section{Experimental results}
\label{sec:experiments}
We apply our methodology to the LeNet-5 on the MNIST dataset, which corresponds to the example provided in \Cref{subsec:case_study}, and other benchmarks: AlexNet on CIFAR-10 (\Cref{subsec:alexnet_cifar10}), AlexNet on CIFAR-100 (\Cref{subsec:alexnet_cifar100}), VGG-16 on CIFAR-10 (\Cref{subsec:vgg_cifar10}) and VGG on CIFAR-100 (\Cref{subsec:vgg_cifar100}). Accuracy improvement results, expressed in terms of Relative Error Reduction, are reported in \Cref{tab:results}. This table also shows the results of Training Time Reduction, power consumption and performance differences between the original and the Hybrid DNN.\\
The algorithm of our methodology (\Cref{alg:methodology}) has been implemented using the pyTorch framework \cite{ref:pytorch} and the experiments have been performed on an Nvidia GTX 1070 GPU (see its specifications in \Cref{tab:GPU_specs}). The power consumptions have been measured using the NVIDIA System Management Interface tool \cite{ref:nvidia_smi}. A schematic view of the experimental setup is shown in \Cref{fig:hw_setup}.\\
\begin{minipage}{\textwidth}
	\begin{minipage}[t]{0.48\linewidth}
		\vspace{11pt}
		\centering
		\includegraphics[width=\linewidth]{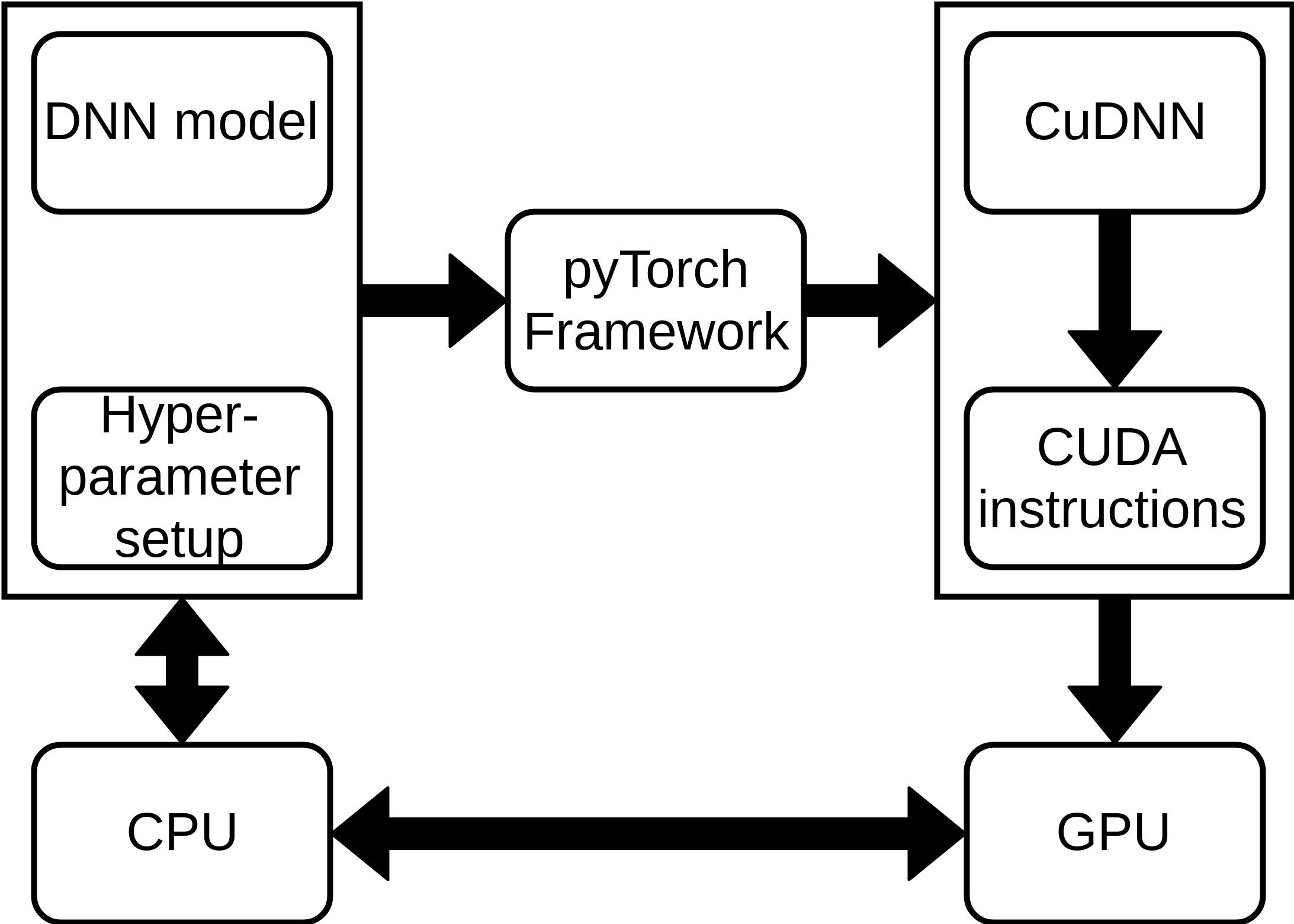}\ \\
		\captionof{figure}{Experimental setup}
		\label{fig:hw_setup}
	\end{minipage}
	\hfill
	\begin{minipage}[t]{0.48\linewidth}
		\vspace{30pt}
		\centering
		\captionof{table}{GPU Specifications}
		\label{tab:GPU_specs}
		%\resizebox{\linewidth}{!}{%
			\begin{tabular}{|c|c|}
				\hline
				\multicolumn{2}{|c|}{\textbf{NVIDIA GTX 1070 specs}} \\ \hline
				CUDA cores & 1920 \\ \hline
				Memory & 8 GB DDR5 \\ \hline
				Mem. interface width & 256-bit \\ \hline
				Mem. bandwidth & 256 GB/s \\ \hline
				Single precision Flops & 6.5 TeraFLOPS \\ \hline
				Power requirement & 150 W \\ \hline
			\end{tabular}%
		%}
	\end{minipage}
\end{minipage}
\begin{table}[h]
	\centering
	\caption{Experimental results, for different benchmarks, in terms of Relative Error Reduction (RER), Evaluation Point (EP), Training Time Reduction (TTR), Power and Performance difference, between the original and the Hybrid DNN.}
	\label{tab:results}
	%\resizebox{\linewidth}{!}{
	\begin{tabular}{|c|c|c|c|c|c|c|}
		\hline
		\textbf{Dataset} & \textbf{Network} & \textbf{RER} & \textbf{EP} & \textbf{TTR} & \textbf{Power} & \textbf{Performance} \\
		\hline
		MNIST & LeNet-5 & 15.56 \% & 7 & 4.29 & + 5.53 \% & - 3.23 \% \\
		\hline
		CIFAR-10 & AlexNet & 7.11 \% & 18 & 7.22 & + 2.37 \% & - 3.24 \% \\
		\hline
		CIFAR-100 & AlexNet & 8.34 \% & 22 & 5.91 & + 1.94 \% & - 3.24 \% \\
		\hline
		CIFAR-10 & VGG-16 & 8.83 \% & 24 & 5.42 & + 3.09 \% & - 3.17 \% \\
		\hline
		CIFAR-100 & VGG-16 & 9.34 \% & 29 & 4.48 & + 2.40 \% & - 3.20 \% \\
		\hline
	\end{tabular}
	%}
\end{table}
\subsection{LeNet-5 on MNIST dataset}
\label{subsec:case_study}
The MNIST benchmark is a collection of handwritten digits (size 28x28), divided into 10 categories. The set consists of 60.000 training images and 10.000 test images. As the original model, we use LeNet-5 architecture \cite{ref:LeNet5}. We analyze the learning curve (\Cref{fig:accuracy_lenet}) and the respective Accuracy Gradient curve (\Cref{fig:accuracy_gradient}) in detail. Looking at the latter curve, we are able to identify the Evaluation Point, which in this case corresponds to the seventh epoch. We then compute the Training Time Reduction (TTR) that we are able to achieve by evaluating the accuracy at the EP instead of at the end of training. Its expression is reported in \Cref{eq:TTR}. The accuracy improvement is measured as the Relative Error Reduction (RER) of the resulting Hybrid DNN with respect to the original one. The RER is defined in \Cref{eq:RER}. Results, including power and performance differences, are reported in the first row of \Cref{tab:results}.\ \\
\begin{equation}
TTR = \frac{\#\ Epochs_{complete\ training}}{EP}
\label{eq:TTR}
\end{equation}
\begin{equation}
RER = \frac{Accuracy_{Hybrid\ model} - Accuracy_{original\ model}}{100 - Accuracy_{original\ model}}
\label{eq:RER}
\end{equation}
\begin{table}[h]
	\begin{varwidth}[b]{0.48\linewidth}
		\centering
		\includegraphics[width=\linewidth]{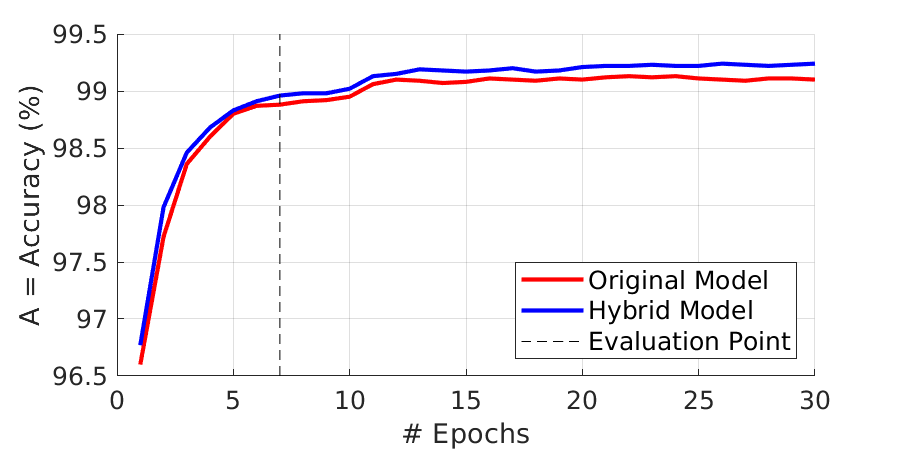}
		\captionof{figure}{Accuracy of the LeNet-5 on the MNIST dataset as a function of the number of epochs.}
		\label{fig:accuracy_lenet}
	\end{varwidth}
	\hfill
	\begin{varwidth}[b]{0.48\linewidth}
		\centering
		\includegraphics[width=\linewidth]{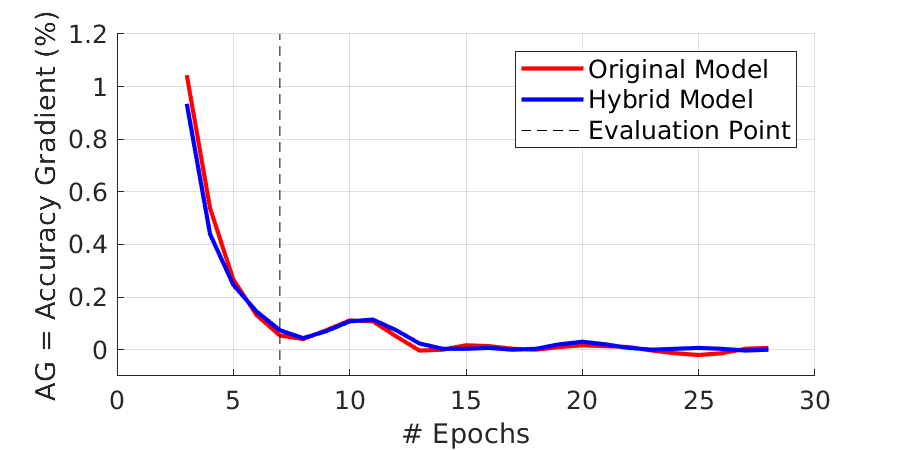}
		\captionof{figure}{Accuracy Gradient of the LeNet-5 on the MNIST dataset as a function of the number of epochs.}
		\label{fig:accuracy_gradient}
	\end{varwidth}
\end{table}
\begin{table}[h]
	\centering
	\caption{Results and activation function configurations of original and Hybrid LeNet-5, trained on the MNIST dataset. Original (first line) and Hybrid (fourth line) models are compared to the network obtained with different Evaluation Points.}
	\label{tab:results_lenet}
	\resizebox{\textwidth}{!}{%
		\begin{tabular}{|c|c|c|c|c|c|c|c|c|c|}
			\hline
			\multirow{2}{*}{\begin{tabular}[c]{@{}c@{}}LeNet-5\\ MNIST\end{tabular}} & \multicolumn{2}{c|}{Layer 1} & \multicolumn{2}{c|}{Layer 2} & \multicolumn{2}{c|}{Layer 3} & Accuracy & RER & TTR \\ \cline{2-10} 
			& ACT & DROP. RATE & ACT & DROP. RATE & ACT & DROP. RATE & - & - & - \\ \hline
			Original & ReLU & 0 & ReLU & 0 & ReLU & 0 & 99.10 \% & - & - \\ \hline
			$EP=5$ & SELU & 0.1 & ReLU & 0.05 & ReLU & 0.1 & 99.17 \% & 7.78 \% & 6x \\ \hline
			$EP=6$ & SELU & 0.1 & ELU & 0.02 & ReLU & 0.05 & 99.21 \% & 12.22 \% & 5x \\ \hline
			\textbf{Hybrid, $EP=7$} & SELU & 0.2 & ELU & 0.02 & ELU & 0.05 & 99.24 \% & 15.56 \% & 4.29x \\ \hline
			$EP=8$ & SELU & 0.2 & ELU & 0.05 & ELU & 0.05 & 99.24 \% & 15.56 \% & 3.75x \\ \hline
			$EP=13$ & SELU & 0.2 & ELU & 0.05 & ELU & 0.1 & 99.26 \% & 17.78 \% & 2.31x \\ \hline
			$EP=30$ & SELU & 0.2 & ELU & 0.05 & ELU & 0.1 & 99.26 \% & 17.78 \% & 1x \\ \hline
		\end{tabular}%
	}
\end{table}\ \\
To prove the efficacy of our methodology, we compare our Hybrid DNN (with versions evaluated at the EP computed with our criteria) with the DNN that is obtained by comparing the results at a different epoch (see \Cref{tab:results_lenet} for the selected activation function and dropout rate combinations for each layer). This analysis leads to a trade-off between accuracy and training speed-up (TTR, Training Time Reduction). \Cref{tab:results_lenet} shows that our EP, corresponding to the seventh epoch, leads to a good solution of the previously discussed trade-off, because an evaluation at a lower epoch significantly reduces the accuracy, while an evaluation at a later epoch reduces the speed-up gain.\\
Moreover, we analyze power and performance differences between the original and the Hybrid DNN, by measuring the relative differences of power and computation time, respectively. The results are reported in \Cref{tab:results}.
\subsection{AlexNet on CIFAR-10 dataset}
\label{subsec:alexnet_cifar10}
The CIFAR-10 dataset \cite{ref:CIFAR} is composed of 50.000 training images and 10.000 test images of size 32x32, divided into 10 different classes. The AlexNet network \cite{ref:AlexNet} consists of 5 convolutional layers and 3 fully-connected layers. Since it was designed to be trained on input images of size 224x224, the first layer of this network has been adapted to the size of CIFAR-10 images. We trained it for 130 epochs using a batch size of 128, with momentum = 0.9 and weight decay = 0.0005. The initial learning rate of 0.1 has been scaled by a factor 0.1 after 40, 80 and 120 epochs. We applied our methodology to this benchmark, obtained a TTR of 7.22x, based on an EP of 18. Our Hybrid DNN achieves 7.11 \% RER, with a power penalty of 1.94 \% and a performance penalty of 3.24 \% with respect to the original AlexNet, as reported in \Cref{tab:results}.
\subsection{AlexNet on CIFAR-100 dataset}
\label{subsec:alexnet_cifar100}
The CIFAR-100 dataset \cite{ref:CIFAR} is composed of 50.000 training images and 10.000 test images of size 32x32, divided into 100 different classes. The DNN, AlexNet, is the same as the one described in \Cref{subsec:alexnet_cifar10} and we trained it with the same hyper-parameters. Our methodology selects an EP of 22, which corresponds to a 8.34 \% TTR. All the other result metrics are reported in \Cref{tab:results}.
\subsection{VGG-16 on CIFAR-10 dataset}
\label{subsec:vgg_cifar10}
The VGG-16 model \cite{ref:VGGNet} is a Deep Neural Network with 13 convolutional layers and 3 fully-connected layers. To comply with input images of size 32x32, a modified version of this network has been used in this experiment. We trained it for the same dataset (CIFAR-10) and the same hyper-parameters as in \Cref{subsec:alexnet_cifar10}. Our methodology produces an EP of 24, which leads to a 5.42 \% TTR. The respective line of \Cref{tab:results} collects the other results.
\subsection{VGG-16 on CIFAR-100 dataset}
\label{subsec:vgg_cifar100}
The same DNN as the one described in \Cref{subsec:vgg_cifar10} (VGG-16) has been trained on CIFAR-100 dataset. For this benchmark, the TTR is 4.48 \%, since the EP is 29. Results are reported in the last row of \Cref{tab:results}.
\section{Conclusions}
\label{sec:conclusions}
In this paper, we have presented an effective methodology for selecting, layer by layer, the type of activation function and the dropout rate associated with it. The performance and power consumptions measured by our Hybrid DNN are slightly worse than the values measured on the original DNN because of the lower complexity of ReLU with respect to ELU and SELU. The accuracy, however, can be improved by a larger factor. Another key contribution of our methodology is the amount of Training Time Reduction in the exploration phase, using the Evaluation Points.
%
%\\RULES OF NIPS:\\
%CAPTION BEFORE THE TABLES!!!\\
%PUT ONE EMPTY ROW BEFORE AND AFTER FIGURES AND TABLES\\
%MAX 8 PAGES + REFERENCES\\
%TODO: REMOVE UNUSED REFERENCES
%\section*{References}

\end{document}